\pdfoutput=1
\documentclass[11pt]{article}

\usepackage{acl}
\usepackage{amsmath}
\usepackage{times}
\usepackage{latexsym}
\usepackage{hyperref}
\usepackage{url}
\usepackage{booktabs}
\usepackage{xcolor}
\usepackage{graphicx}
\usepackage{subcaption}
\usepackage{soul}
\usepackage[T1]{fontenc}
\usepackage[utf8]{inputenc}
\usepackage{microtype}

\title{The Effect of Language Diversity When Fine-Tuning\\ Large Language Models for Translation}

 \author{
    {\bf David Stap\hspace{2mm}}
    {\bf Christof Monz\hspace{2mm}}\\
    Language Technology Lab\\
    University of Amsterdam\\
    \medskip
    \texttt{\{d.stap, c.monz\}@uva.nl}
 }

\begin{document}

\maketitle

\begin{abstract}
Prior research diverges on language diversity in LLM fine-tuning: Some studies report benefits while others find no advantages.
Through controlled fine-tuning experiments across 132 translation directions, we systematically resolve these disparities.
We find that expanding language diversity during fine-tuning improves translation quality for both unsupervised and---surprisingly---supervised pairs, despite less diverse models being fine-tuned exclusively on these supervised pairs.
However, benefits plateau or decrease beyond a certain diversity threshold.
We show that increased language diversity creates more language-agnostic representations.
These representational adaptations help explain the improved performance in models fine-tuned with greater diversity.
\end{abstract}

\section{Introduction}
\label{sec:introduction}
General-purpose LLMs like \textsc{Llama~3} \citep{grattafiori_llama_2024} show promise for machine translation but require targeted fine-tuning beyond their incidental bilingualism \citep{briakou_searching_2023} to match the performance of specialized translation systems.
Through fine-tuning approaches ranging from two-stage methods \citep{li_eliciting_2023,zeng_teaching_2024} to more sophisticated optimization techniques \citep{xu_contrastive_2025,zhu-etal-2024-preference}, LLMs such as \textsc{Tower} \citep{alves_tower_2024} now outperform traditional NMT systems \citep{kocmi-etal-2024-findings,deutsch_wmt24_2025}.

Current research presents conflicting evidence on multilingual fine-tuning strategies. 
Some studies show that scaling the number of tasks or languages during instruction tuning improves (cross-lingual) generalization \citep{wang_super-naturalinstructions_2022, muennighoff-etal-2023-crosslingual, dang-etal-2024-rlhf}, while others report that just 1--3 fine-tuning languages effectively trigger cross-lingual transfer \citep{kew_turning_2024,zhu_finetuning_2024}.
Recent \textit{inference-only} experiments by \citet{richburg_how_2024} across 132 translation directions highlight this uncertainty, showing variance in translation quality with off-target generations for non-English sources and inconsistent performance across languages.
While non-English over-tokenization and typological distance provide partial explanations, controlled fine-tuning experiments on the effects of language diversity \textit{during fine-tuning} remain unexplored.

We address these conflicting findings through systematic experimentation with varying translation directions, measuring effects on both seen and unseen language pairs.
Through controlled fine-tuning across 132 translation directions, we demonstrate that increasing language diversity consistently improves translation quality in all categories.
Counterintuitively, models fine-tuned on the most diverse language sets outperform others \textit{even on fully supervised language pairs} that less diverse models are specifically optimized to handle.
However, experiments with even larger language sets (272 directions) reveal that benefits plateau or decrease beyond a certain diversity threshold.
Analysis of model activations shows that fine-tuning on diverse language directions creates more target language-agnostic representations in middle layers, explaining the performance improvements in our most diverse models.

\begin{figure*}[ht!]
    \centering
    \includegraphics[width=.495\linewidth]{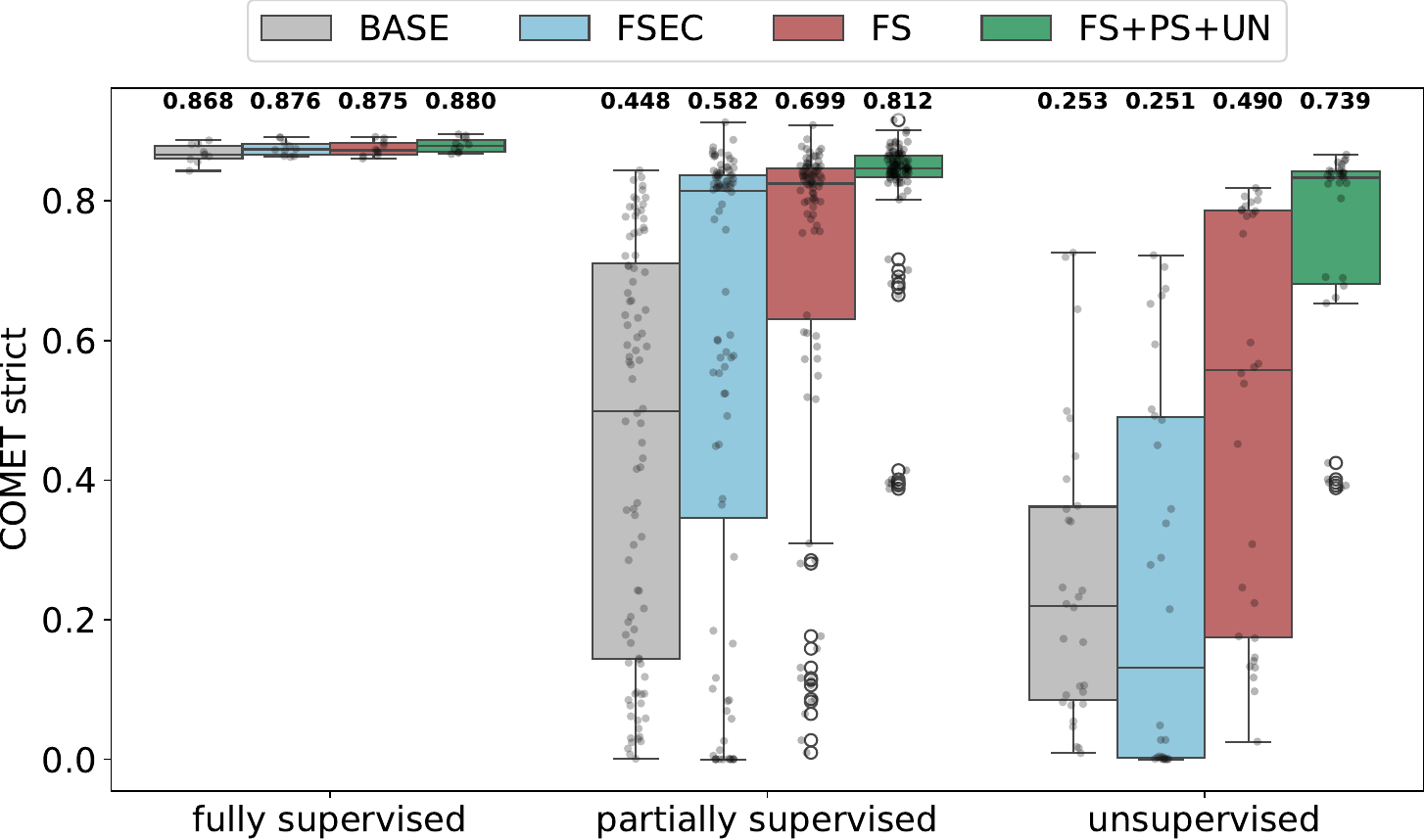}
    \includegraphics[width=.495\linewidth]{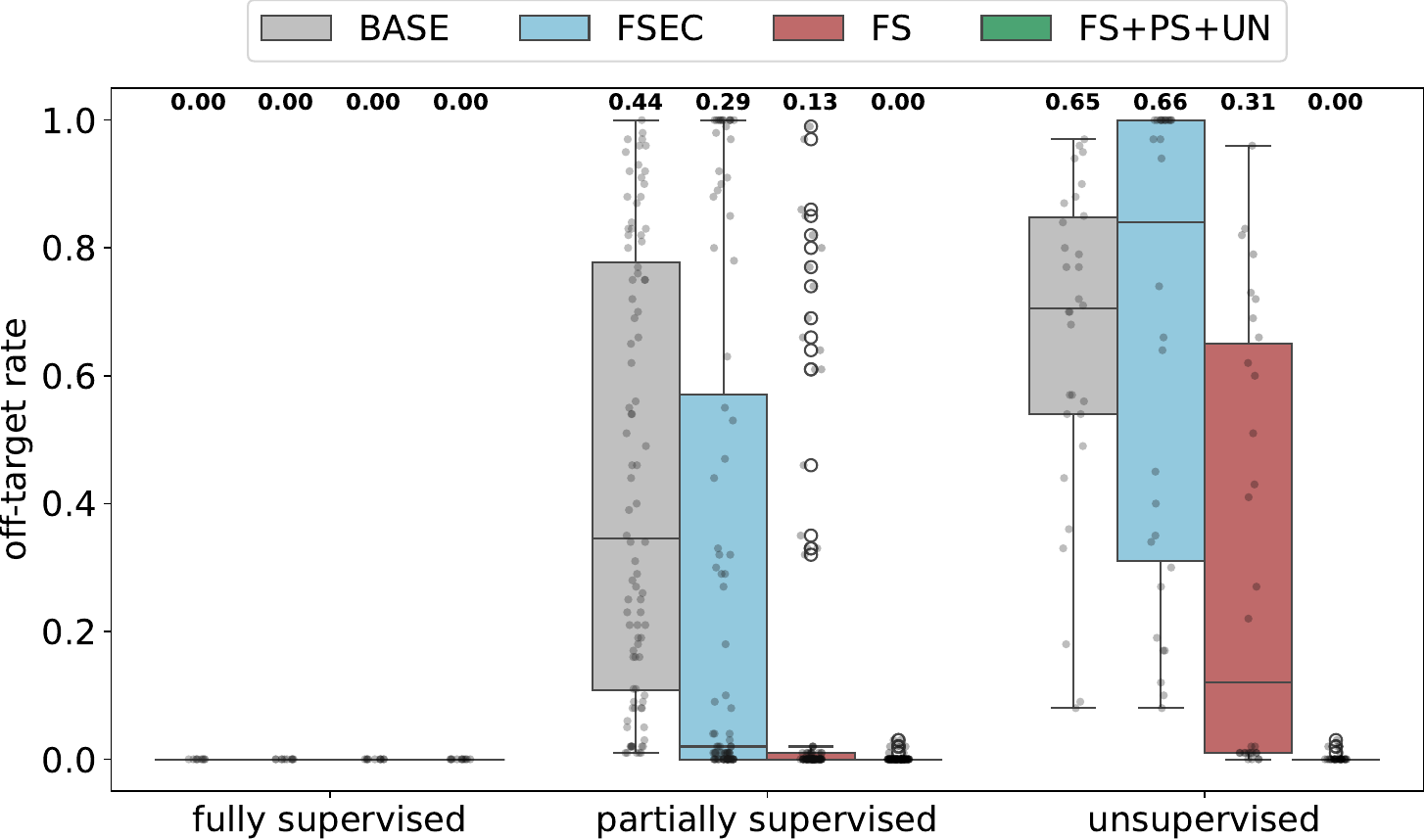}
    \caption{\small{\textsc{COMET-strict} scores (left) and off-target rates (right) for \textsc{base} (no fine-tuning), \textsc{fsec} (English-centric), \textsc{fs} (seen directions), \textsc{fs+ps+un} (all directions), evaluated on \textit{fully supervised} (\texttt{de}/\texttt{en}/\texttt{ko}/\texttt{nl}/\texttt{ru}/\texttt{zh} pairs), \textit{unsupervised} (\texttt{cs}/\texttt{is}/\texttt{ja}/\texttt{pl}/\texttt{sv}/\texttt{uk} pairs), and \textit{partially supervised} (combining supervised and unsupervised) language pairs.
    Numbers above bars show mean scores.
    Training on more diverse sets improves \textit{all} categories, with \textsc{fs+ps+un} achieving best \textsc{COMET-strict} scores even for fully supervised pairs.
    \textsc{fs} substantially reduces off-target rates for unsupervised directions compared to \textsc{base} and \textsc{fsec}, despite these pairs being \textit{absent} from its fine-tuning data.}} 
    \label{fig:comet-strict-off-target-7b}
\end{figure*}

\section{Fine-tuning and evaluation design}
Following \citet{richburg_how_2024}, we categorize our language pairs into three groups based on their presence in the fine-tuning data of the \textsc{Tower} model we build upon: \textit{fully supervised} (pairs between \texttt{de}, \texttt{en}, \texttt{ko}, \texttt{nl}, \texttt{ru} and \texttt{zh}), \textit{zero-shot} (pairs involving \texttt{cs}, \texttt{is}, \texttt{ja}, \texttt{pl}, \texttt{sv} and \texttt{uk}), and \textit{partially supervised} (pairs combining supervised and zero-shot languages).
This yields 132 translation directions across 12 typologically diverse languages with varying pre-training representation, enabling comprehensive assessment across different data conditions (see Appendix~\ref{app:lang_details}).

\paragraph{Fine-tuning setups}
We compare the following incremental fine-tuning approaches using the \textsc{Tower} family of models, which are built on \textsc{Llama 2} and underwent continued pre-training with a mixture of monolingual and parallel data:

\begin{itemize}
    \item \textbf{\textsc{base}}: \textsc{TowerBase-7B} model without task-specific fine-tuning, serving as our baseline.
    \textbf{\textsc{fsec}}: \textsc{base} fine-tuned only on fully supervised English-centric translation directions (10 directions), representing minimal supervision.
    \item \textbf{\textsc{fs}}: \textsc{base} fine-tuned on all fully supervised language directions (30 directions), extending beyond English-centric pairs to investigate transfer learning between diverse language combinations.
    \item \textbf{\textsc{fs+ps+un}}: \textsc{base} fine-tuned on fully supervised, partially supervised, and unsupervised directions (132 directions), maximizing language diversity to investigate cross-lingual transfer effects.
\end{itemize}

This controlled experimental design allows us to systematically evaluate how increasing language diversity during fine-tuning affects both supervised and unsupervised translation directions, moving beyond aggregate scores to understand performance patterns across specific language groups.

\paragraph{Data}
We fine-tune on \textsc{NTREX-128} \citep{federmann_ntrex-128_2022}, a high-quality dataset of 1,997 multi-parallel professionally translated sentences designed for machine translation evaluation.\footnote{Preliminary experiments with additional \textsc{FLORES-200} (\texttt{dev}) data showed no significant improvements, so we exclude it for experimental clarity.}
For evaluation, we use the \textsc{FLORES-200} \citep{nllb_team_no_2022} \texttt{devtest} set, which provides multi-parallel data for controlled cross-language comparison.

\paragraph{Metrics}
Our primary metric is \textsc{COMET-strict}, a modified version of \textsc{COMET} \citep{rei_comet_2020} that assigns zero scores to off-target translations, following recommendations by \citet{zouhar_pitfalls_2024}.\footnote{We use version \texttt{wmt22-comet-da}.}
We also report off-target rates, measured using \textsc{fastText} \citep{joulin-etal-2017-bag,joulin_fasttextzip_2016} language identification.\footnote{We use the \texttt{lid.176.bin} model.}
Optimization and inference details are provided in Appendix~\ref{app:impl}.

\section{Results}

\paragraph{Increased diversity leads to better performance}

Figure~\ref{fig:comet-strict-off-target-7b} (left) demonstrates that expanding language diversity during fine-tuning yields consistent performance improvements across all language pair categories.
The \textsc{COMET-strict} scores show a clear progression from \textsc{base} to \textsc{fsec} to \textsc{fs} to \textsc{fs+ps+un} models, with the most diverse model achieving the highest scores in every category.
Surprisingly, the \textsc{fs+ps+un} model (fine-tuned on \textit{all} 132 directions) outperforms specialized models even on fully supervised language pairs (0.880 vs.\ 0.876 for \textsc{fsec}), despite the latter being specifically optimized for these directions.
The benefits become more pronounced for partially supervised (0.812 vs.\ 0.448 for \textsc{base}) and unsupervised (0.739 vs.\ 0.253 for \textsc{base}), although this improvement is expected as \textsc{fs+ps+un} is explicitly fine-tuned on these directions.

These results clarify conflicting evidence on language diversity (see §\ref{sec:introduction}) and align with \citet{wang_super-naturalinstructions_2022} and \citet{dang-etal-2024-rlhf}, confirming that \textit{broad language diversity} (132 directions vs. 10--30), rather than minimal exposure, substantially enhances cross-lingual transfer, even for pairs already well supported in more specialized models.

\paragraph{Increased diversity reduces off-target problem}
Off-target translations, where models generate content in incorrect languages, represent a critical failure mode in LLM-based MT \citep{zhang_prompting_2023,guerreiro_hallucinations_2023,sennrich-etal-2024-mitigating}.

Figure~\ref{fig:comet-strict-off-target-7b} (right) shows that while all models maintain target language fidelity for fully supervised pairs, the \textsc{base} model produces incorrect target languages at alarming rates for partially supervised (44\%) and unsupervised pairs (65\%).
Fine-tuning progressively mitigates this problem, with \textsc{fs} showing substantial improvements (13\% and 31\% respectively) despite not being explicitly trained on these language combinations.
Significantly, the \textsc{fs+ps+un} model completely eliminates off-target translations across all categories.

\begin{figure}[t]
    \centering
    \includegraphics[width=\linewidth]{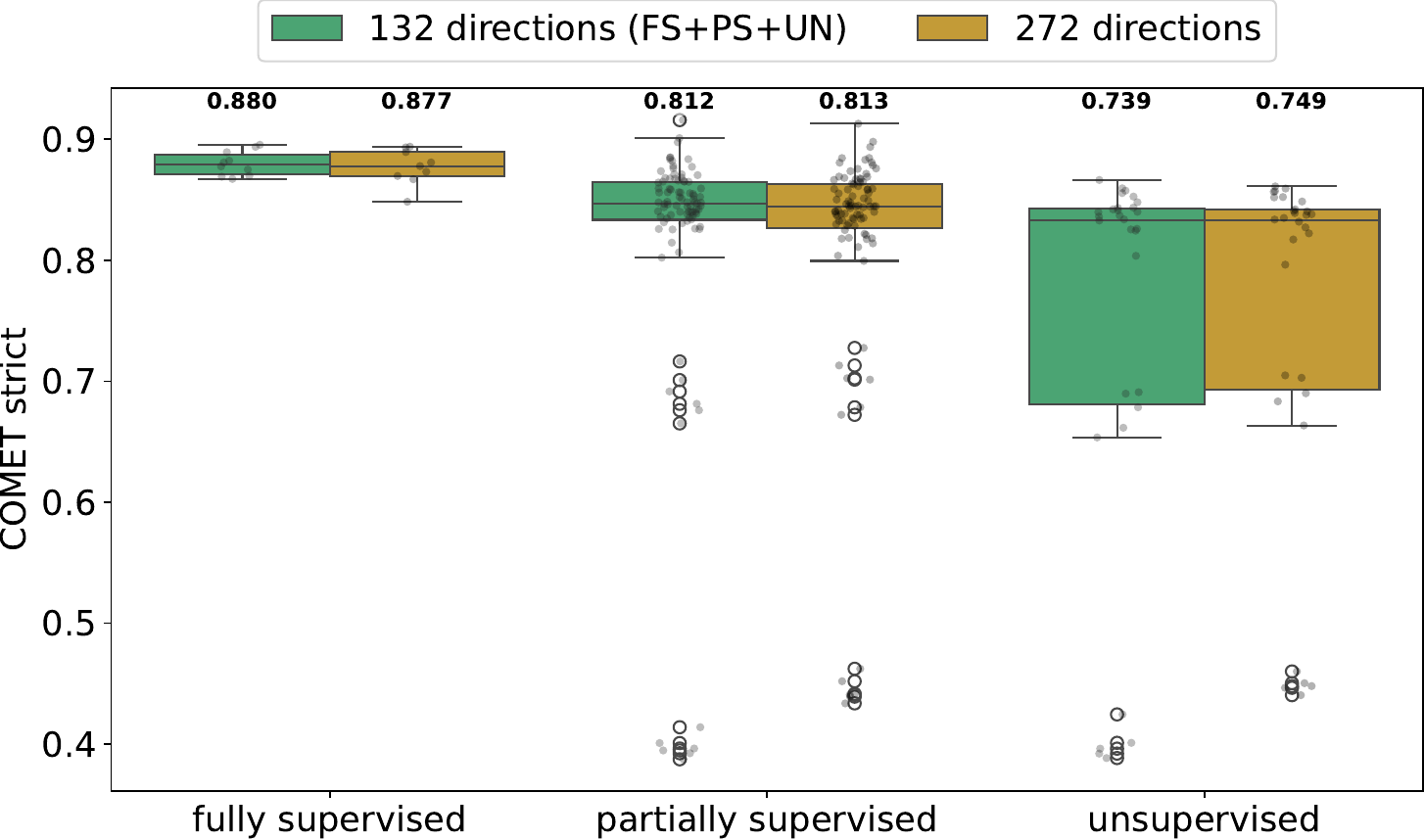}
    \caption{\small{\textsc{COMET-strict} scores comparing models trained on 132 directions and 272 directions.
    Both are evaluated on the original test set with the same language pairs as used throughout the paper.
    Unsupervised directions show the clearest benefits from increased diversity (+0.01), while fully supervised directions show a slight decrease (-0.003), suggesting potential diversity trade-offs.}}
    \label{fig:comet-strict-7b-272}
\end{figure}

\paragraph{Diversity benefits plateau}
To probe the limits of linguistic breadth, we expanded 	extsc{fs+ps+un} from 132 to 272 directions by adding Danish, Afrikaans, Slovak, Bulgarian, and Vietnamese while preserving the family balance of our original design.
These languages appear only during fine-tuning—the evaluation set remains identical to isolate the effect of additional training diversity.
Figure~\ref{fig:comet-strict-7b-272} contrasts the two setups.
Unsupervised directions gain the most (+0.01 \textsc{COMET-strict}), partially supervised pairs remain stable (+0.001), and fully supervised pairs dip slightly (-0.003), revealing a plateau for well-supported languages.
This pattern refines earlier reports of monotonic gains with diversity \citep{wang_super-naturalinstructions_2022,dang-etal-2024-rlhf} and echoes the diminishing returns seen in multilingual pretraining \citep{muennighoff-etal-2023-crosslingual}.

\paragraph{Regularization alone insufficient}
Regularization benefits models by enhancing generalization and calibration, with strong effects when using distant languages \citep{meng-monz-2024-disentangling}.
We investigate whether these benefits can be achieved through explicit regularization techniques (weight decay and LoRA) rather than language diversity, but find no comparable improvements. 
This aligns with \citet{aharoni_massively_2019}, who suggest that multilingualism provides benefits beyond explicit regularization methods.
See Appendix~\ref{app:regularization} for details.

\paragraph{Results not due to multi-parallel data}
While recent work by \citet{caswell_smol_2025} found that fine-tuning on multi-parallel data causes catastrophic forgetting in LLMs when trained on X$\rightarrow$\texttt{en} directions, our findings persist beyond multi-parallel settings.
We replicated our experiments using non-multi-parallel data scraped from OPUS and observed similar diversity benefits (see Appendix~\ref{app:multi_parallel}).
Unlike the overfitting issues reported for LLMs, our models maintain performance, consistent with prior work showing multi-parallel data benefits in NMT \citep{stap_viewing_2023,wu-etal-2024-far}.

\paragraph{Findings persist at larger scale}
Larger models (13B) exhibit the same trends: increased language diversity leads to reduced off-target rates and improved cross-lingual transfer.
This confirms our findings are robust across model scales.
Complete experimental details are provided in Appendix~\ref{app:scale_invariance}.

\paragraph{Findings transfer across architectures}
We further replicate our setups on the multilingual \textsc{Gemma~2~2B} model \citep{team_gemma_2025} to test whether diversity benefits extend beyond the \textsc{Tower} family. Table~\ref{tab:gemma2} shows the same monotonic improvements we observe for \textsc{Tower}: expanding the fine-tuning directions from \textsc{fsec} to \textsc{fs} to \textsc{fs+ps+un} consistently raises \textsc{COMET-strict} scores across supervision regimes. The smaller architecture starts from stronger zero-shot performance than \textsc{TowerBase} (0.469 vs. 0.253 on unsupervised pairs) yet still gains the most from the full 132-direction setup (\textsc{fs+ps+un}), confirming that language-diversity gains are not model-family artifacts.

\begin{table}[t]
    \centering
    \small
    \begin{tabular}{lccc}
    \toprule
    Model & FS & PS & UN \\
    \midrule
    \textsc{base} & 0.620 & 0.522 & 0.469 \\
    \textsc{fsec} & 0.723 & 0.618 & 0.498 \\
    \textsc{fs} & 0.725 & 0.645 & 0.587 \\
    \textsc{fs+ps+un} & 0.739 & 0.711 & 0.697 \\
    \bottomrule
    \end{tabular}
    \caption{\small{\textsc{COMET-strict} scores for fine-tuning \textsc{Gemma~2~2B} across the supervision regimes.
    Columns FS, PS, and UN correspond to fully supervised, partially supervised, and unsupervised pairs.
    Language-diversity progression mirrors the trends observed for \textsc{Tower}, demonstrating cross-architecture robustness.}}
    \label{tab:gemma2}
\end{table}

\paragraph{Middle layers adapt most}

\begin{figure}[ht]
    \centering
    \includegraphics[width=\linewidth]{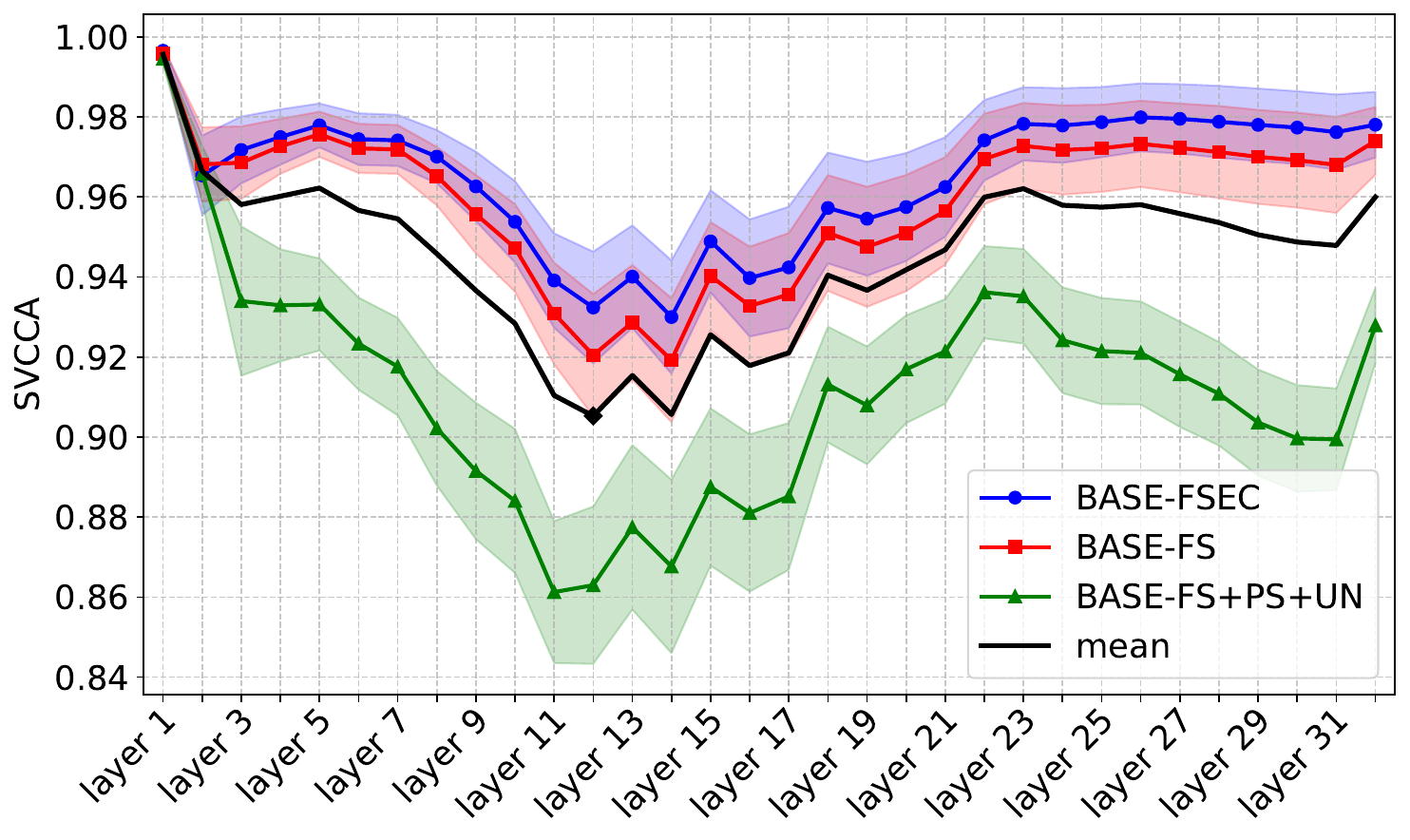}
    \caption{\small{SVCCA similarity scores between fine-tuned and \textsc{base} models across layers. 
    Lower values indicate greater adaptation during fine-tuning. 
    \textsc{base-fsec} (blue), \textsc{base-fs} (red), and \textsc{base-fs+ps+un} (green) are compared, with their mean shown in black. 
    Shaded regions represent confidence intervals.
    Middle layers show most significant adaptation, with lowest mean similarity (0.91) at layer 12. 
    \textsc{fs+ps+un} exhibits greater adaptation throughout the network.}}
    \label{fig:svcca_per_layer}
\end{figure}

We analyze activation patterns across models by comparing them with the base model using Singular Vector Canonical Correlation Analysis (SVCCA; \citealp{raghu_svcca_2017}).
This analysis identifies \textit{where} and \textit{to what extent} adaptations occur during fine-tuning.
We aggregate activations across all source-target language pairs and present the layer-specific results in Figure \ref{fig:svcca_per_layer}.

Our analysis reveals that middle layers consistently undergo the most substantial adaptation across all fine-tuned models, with the lowest mean similarity (0.91) occurring at layer 12.
Furthermore, models fine-tuned on more languages exhibit greater divergence from the base model, with \textsc{fs+ps+un} showing most substantial adaptations.

Middle layers encode semantic information and show the strongest cross-lingual transfer capabilities \citep{liu_middle-layer_2025,liu-etal-2025-selected}.
Our findings support that larger degrees of cross-lingual transfer within middle layers explain the performance improvements observed in models fine-tuned on a larger linguistic diversity.

\paragraph{Diversity increases cross-lingual overlap}

\begin{figure}[t!]
    \centering
    \includegraphics[width=\linewidth]{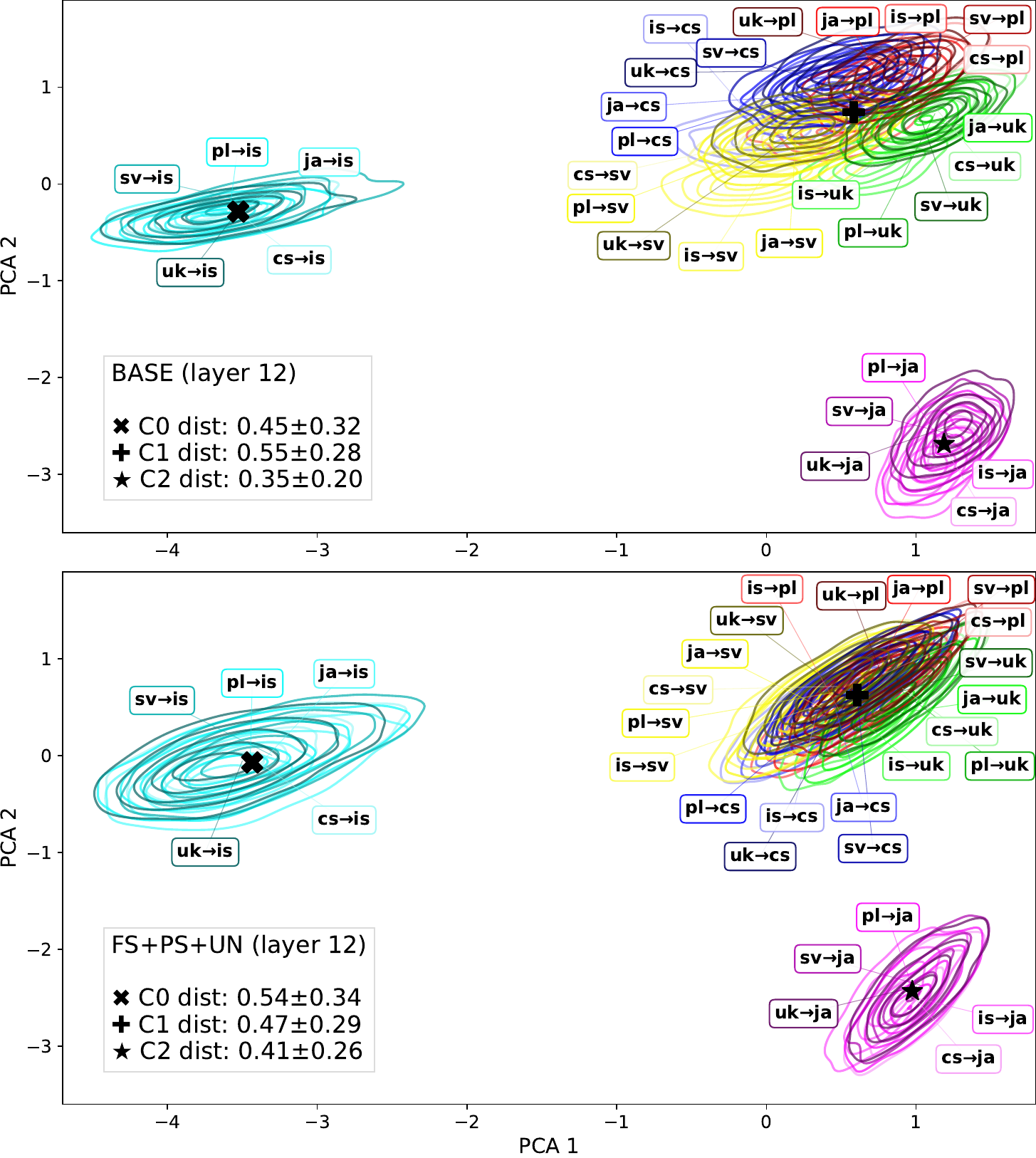}
    \caption{\small{Kernel density estimation of layer 12 activations for \textsc{base} (top) and \textsc{fs+ps+un} (bottom).
    Colors represent translation directions; markers denote k-means clusters: $\times$ (\texttt{is}$\rightarrow$\texttt{en}), $+$ (\texttt{cs}/\texttt{pl}/\texttt{sv}/\texttt{uk}$\rightarrow$\texttt{en}), and $\star$ (\texttt{ja}$\rightarrow$\texttt{en}).
    Intra-cluster distances show increased specialization for single-target clusters in \textsc{fs+ps+un}, while multi-target cluster C1 demonstrates increased overlap.}}
    \label{fig:kde_layer_12}
\end{figure}

We analyze layer 12 (the most significantly modified layer) to understand \textit{which adaptations} occur during fine-tuning.
Following from \citet{gao_towards_2024} and \citet{wang_bridging_2024}, we apply t-SNE dimension reduction \citep{vandermaaten2008visualizing} to layer activations and visualize the bivariate kernel density (KDE) estimation.
Next, we employ \textit{k}-means clustering to identify language groups within these representations, using silhouette score maximization \citep{ROUSSEEUW198753} for optimal cluster determination without requiring manual inspection.
Finally, we calculate the intra-cluster distances.
We compare the \textsc{base} and \textsc{fs+ps+un} models, visualizing unsupervised directions where we expect the most significant adaptations.

Figure \ref{fig:kde_layer_12} presents the resulting visualization.
Notably, for the single-target language clusters C0 and C2, the \textsc{fs+ps+un} model exhibits \textit{greater intra-cluster distances} (0.54±0.34 and 0.41±0.26) compared to the \textsc{base} model (0.45±0.32 and 0.35±0.20), suggesting \textit{increased specialization} per source-target direction after fine-tuning on diverse data.
Conversely, for the multi-target language cluster (C1), the \textsc{fs+ps+un} model shows \textit{reduced} intra-cluster distances (0.47±0.29) relative to the \textsc{base} model (0.55±0.28), indicating \textit{greater representational overlap} between these linguistically related languages.
This increased overlap provides evidence for enhanced cross-lingual transfer, which contributes to the superior performance of models fine-tuned on greater linguistic diversity.

Table \ref{tab:cluster_distances} presents intra-cluster distances for all models.
Note that clusters contain the same languages for all setups.
As diversity increases, single-target clusters (C0, C2) show greater specialization while multi-language cluster C1 exhibits enhanced representational overlap, suggesting improved cross-lingual transfer.

While previous work has \textit{explicitly} aligned representations \citep{liu_middle-layer_2025,kargaran_mexa_2024,stap_viewing_2023}, our findings show \textit{implicit} alignment occurs through multilingual fine-tuning.

\begin{table}[htbp]
    \centering
    \small
    \begin{tabular}{lccc}
    \toprule
    &  $\boldsymbol{\times}$ C0 & $\boldsymbol{+}$ C1 & $\boldsymbol{\star}$ C2 \\
    \midrule
    \textsc{base}        & $0.45\pm0.32$ & $0.55\pm0.28$ & $0.35\pm0.20$\\
    \textsc{fsec}        & $0.49\pm0.33$ & $0.53\pm0.26$ & $0.34\pm0.20$\\
    \textsc{fs}          & $0.52\pm0.36$ & $0.51\pm0.28$ & $0.39\pm0.24$\\
    \textsc{fs+ps+un}    & $0.54\pm0.34$ & $0.47\pm0.29$ & $0.41\pm0.26$\\
    \bottomrule
    \end{tabular}
    \caption{\small{Intra-cluster distances.
    C0 (\texttt{is} target) and C2 (\texttt{ja} target) show increased distances in models fine-tuned on more diverse data, while C1 (\texttt{cs}, \texttt{pl}, \texttt{sv}, \texttt{uk} targets) shows decreased distances, indicating enhanced cross-lingual transfer.}}
    \label{tab:cluster_distances}
\end{table}

\section{Conclusion}
Our systematic investigation across 132 translation directions resolves conflicting findings on language diversity in LLM fine-tuning.
We show that fine-tuning on broader language sets consistently improves translation across all categories: fully supervised, partially supervised, and zero-shot pairs.
Consequently, we recommend fine-tuning with diverse language directions even when optimizing for a limited subset of target translation pairs, as our most diverse model outperformed models specialized exclusively for those target pairs.
However, we advise identifying an optimal diversity threshold, as too many languages diminishes performance for well-supported pairs while still benefiting less-represented languages.
Our representational analysis attributes the diversity improvements to specific adaptations in middle layers, revealing increased language-agnostic representations, which explains the enhanced cross-lingual transfer.

\section*{Limitations}
We evaluate on the \textsc{FLORES-200} \citep{nllb_team_no_2022} \texttt{devtest} set, a multi-parallel benchmark consisting of documents originally written in English and professionally translated into multiple languages.
While this may introduce some translationese effects, the multi-parallel nature enables controlled comparison across language pairs.

Our findings are based on the \textsc{Tower} model family \citep{alves_tower_2024} (7B and 13B), built on \textsc{LLaMA 2} \citep{touvron_llama2_2023}.
Further research should verify whether these patterns generalize to other model architectures and even larger model sizes.

\section*{Acknowledgments} 
This research was funded in part by the Netherlands Organization for Scientific Research (NWO) under project number VI.C.192.080.
We thank SURF (www.surf.nl) for the support in using the National Supercomputer Snellius.

\bibliography{references}
\bibliographystyle{acl_natbib}

\appendix

\begin{table*}[ht]
    \begin{center}
        \small
        \begin{tabular}{lllll}
        \toprule
        \textbf{Language} & \textbf{ISO 639-1} & \textbf{Script} & \textbf{LLaMA 2 support} & \textbf{Similarity groups} \\
        \midrule
        Czech            & \texttt{cs} & Latin    & 0.03\%          & West Slavic \\
        Polish           & \texttt{pl} & Latin    & 0.09\%          & West Slavic \\
        \textbf{Russian} & \texttt{ru} & Cyrillic & 0.13\%          & East Slavic \\
        Ukrainian        & \texttt{uk} & Cyrillic & 0.07\%          & East Slavic \\\midrule
        \textbf{German}  & \texttt{de} & Latin    & 0.17\%          & West Germanic \\
        \textbf{English} & \texttt{en} & Latin    & 89.70\%         & West Germanic \\
        Icelandic        & \texttt{is} & Latin    & possibly unseen & North Germanic \\
        \textbf{Dutch}   & \texttt{nl} & Latin    & 0.12\%          & West Germanic \\
        Swedish          & \texttt{sv} & Latin    & 0.15\%          & North Germanic \\\midrule
        Japanese         & \texttt{ja} & Kana     & 0.10\%          & Kanji from Hanzi, SOV order \\
        \textbf{Korean}  & \texttt{ko} & Hangul   & 0.06\%          & SOV order \\
        \textbf{Chinese} & \texttt{zh} & Hanzi    & 0.13\%          & Hanzi to Kanji, loanwords to \texttt{ja} and \texttt{ko} \\
        \bottomrule
        \end{tabular}
    \end{center}
    \caption{Evaluated languages with rationales for similarity grouping, following the language selection from \citet{richburg_how_2024}.
    Languages marked in \textbf{bold} belong to the supervised set used in the original \textsc{Tower} model fine-tuning.}
    \label{tab:lang_details}
\end{table*}

\begin{table*}[ht]
    \begin{center}
        \small
        \begin{tabular}{l}
        \toprule
        \texttt{Translate this from \{source\_language\} to \{target\_language\}:} \\
        \texttt{\{source\_language\}: \{source\_sentence\}} \\
        \texttt{\{target\_language\}: \{target\_sentence\}} \\
        \bottomrule
        \end{tabular}
    \end{center}
    \caption{Prompting template for fine-tuning and 0-shot inference.
    For fine-tuning \texttt{\{target\_sentence\}} is filled with the corresponding target sentence, and for 0-shot inference it is the empty string.}
    \label{tab:prompt}
\end{table*}

\section{Language details}
\label{app:lang_details}
The selection of languages shown in Table \ref{tab:lang_details}, following the language selection from \citet{richburg_how_2024}, enables evaluation across varied typological properties and scripts while providing a systematic comparison between supervised languages (seen during fine-tuning) and zero-shot languages that share linguistic features with the supervised set.
The languages in the zero-shot set were chosen to represent both varying degrees of resource support in the pre-training data and to have relationships to languages in the supervised set through language family, typological properties, or orthography.

\section{Implementation details}
\label{app:impl}

\subsection{Optimization}
We conducted hyperparameter tuning on our development set (\textsc{FLORES-200} \texttt{dev}), exploring learning rate scheduler $\in\{\text{cosine, inverse square root}\}$, batch size $\in\{128, 256\}$, and learning rate $\in\{2 \times 10^{-5}, 2 \times 10^{-6}\}$.

For all experiments, we performed full fine-tuning using the AdamW optimizer \citep{loshchilov2018decoupled} with 5\% warm-up percentage and trained for one epoch.
Based on development set performance, we selected the optimal configuration: a cosine learning rate scheduler with batch size of 256 and learning rate of $2 \times 10^{-5}$.
We implemented our fine-tuning experiments using the Hugging Face transformers library \citep{wolf_transformers_2020} with DeepSpeed \citep{rasley_deepspeed_2020}.

\subsection{Inference}
For both fine-tuning and zero-shot inference, we used the prompt template shown in Table~\ref{tab:prompt}.
We mask out the prompt during fine-tuning.
We employed greedy decoding (beam size 1) to balance computational efficiency with comprehensive evaluation across all translation directions.

\section{Additional results}
\label{app:additional_results}

\subsection{Regularization alone insufficient}
\label{app:regularization}

\begin{figure}[ht]
    \centering
    \includegraphics[width=\linewidth]{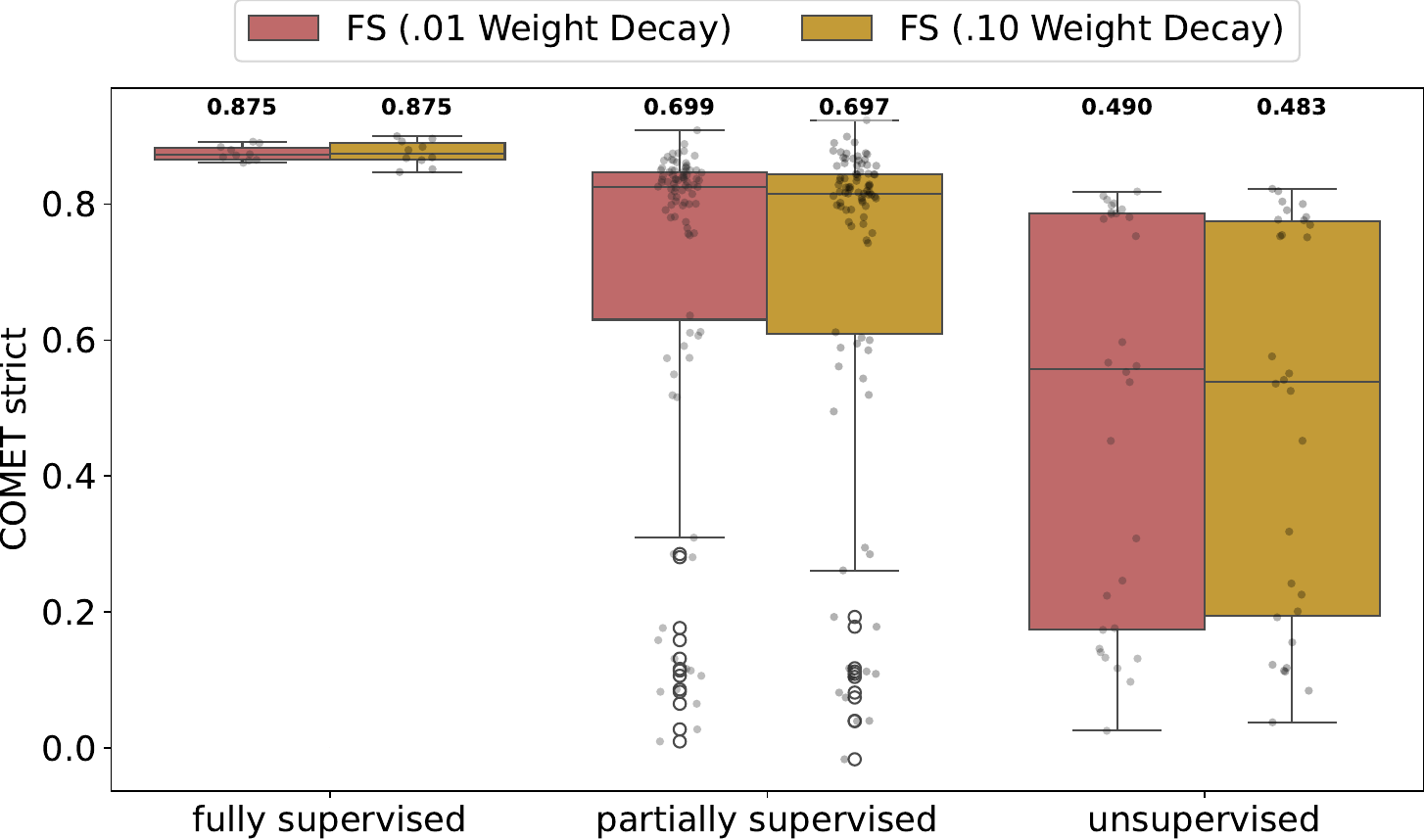}
    \caption{\small{\textsc{COMET-strict} scores comparing \textsc{fs} models with weight decay values of 0.01 (standard) and 0.10.
    Increasing regularization strength shows minimal impact on fully supervised and partially supervised directions, while actually harming performance on unsupervised directions, suggesting that regularization alone cannot replicate the benefits of increased language diversity.}}
    \label{fig:regularization}
\end{figure}

\begin{figure*}[ht]
    \centering
    \includegraphics[width=.495\linewidth]{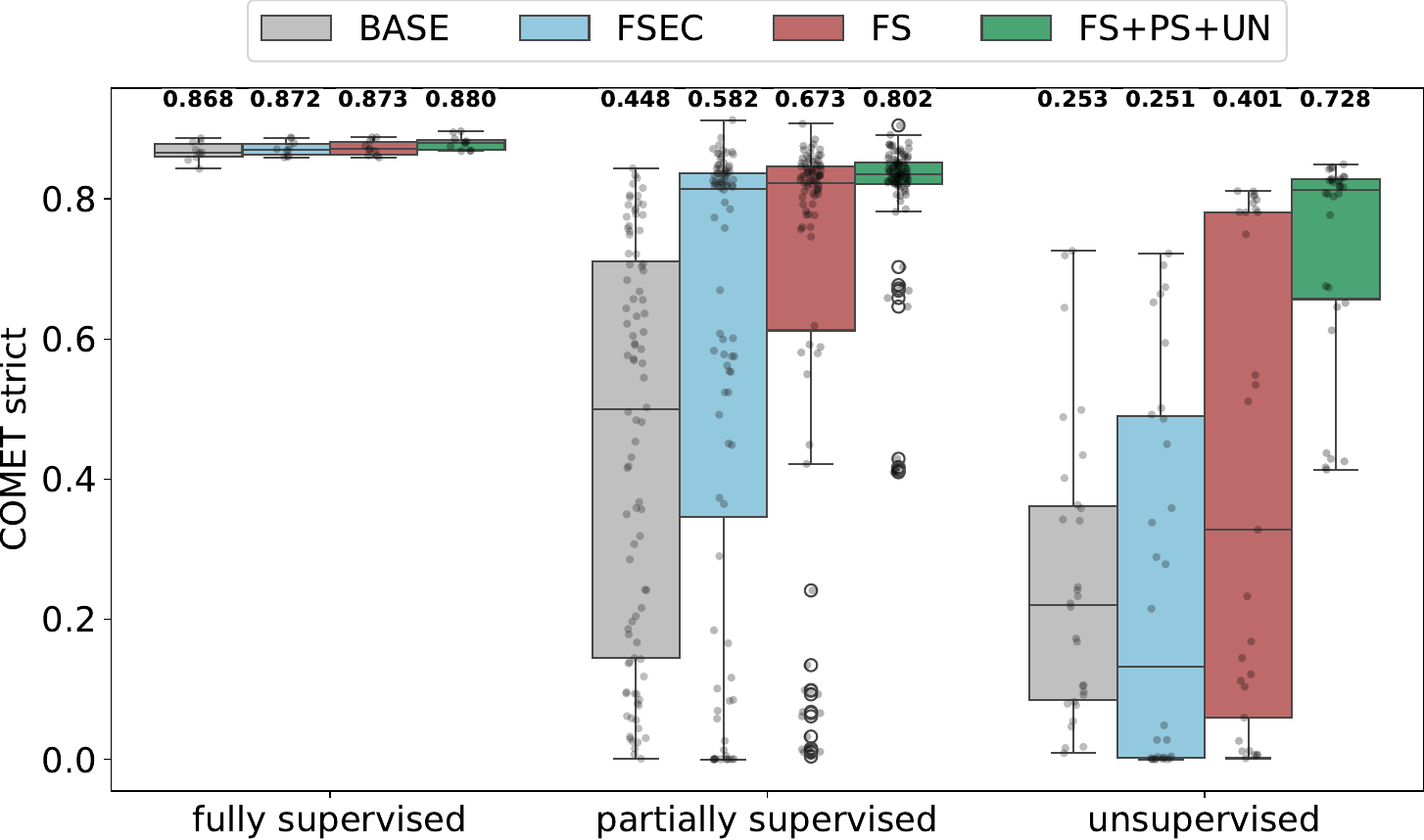}
    \includegraphics[width=.495\linewidth]{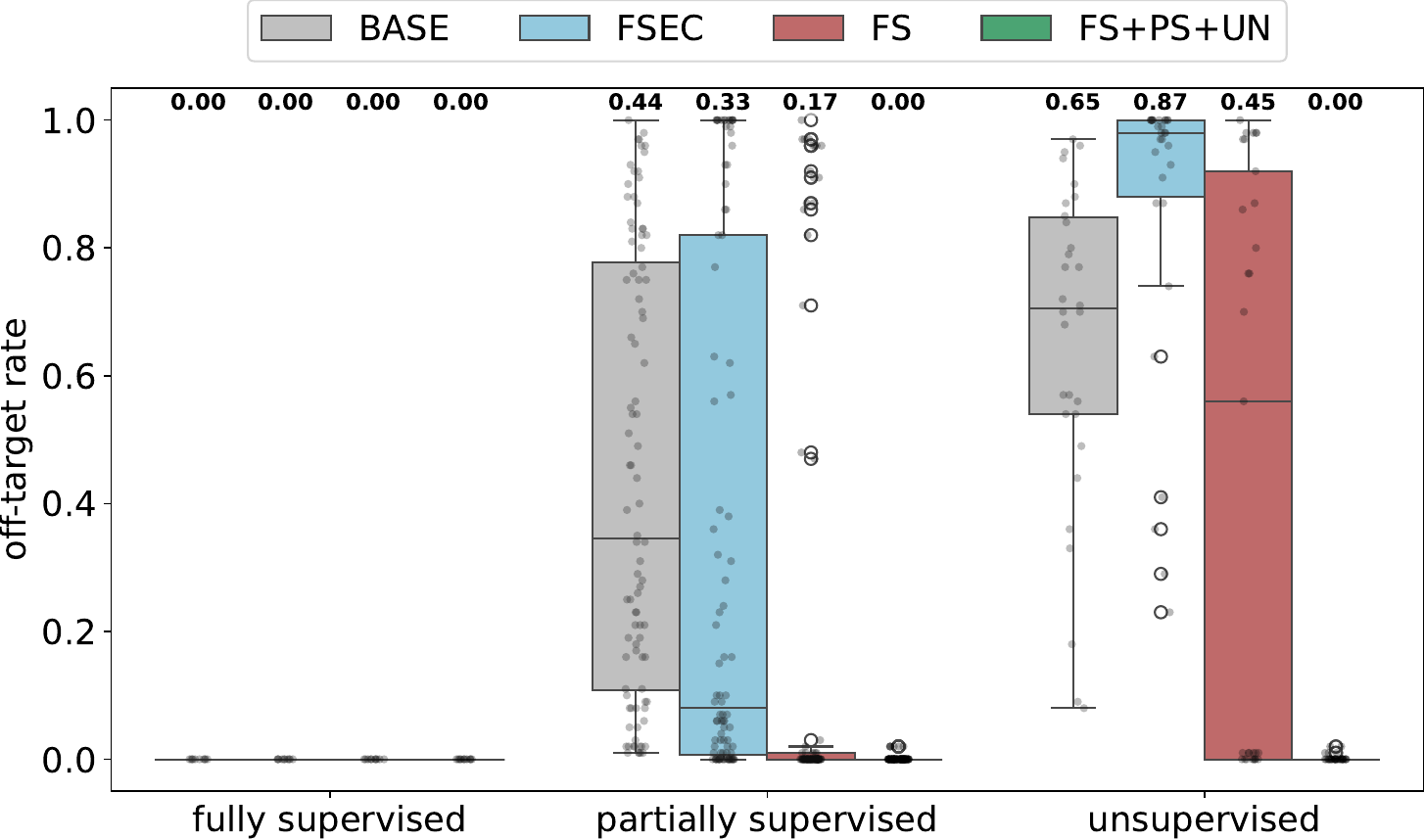}
    \caption{\small{\textsc{COMET-strict} scores for 7B models fine-tuned on filtered \textsc{NLLB} dataset: \textsc{base} (no fine-tuning), \textsc{fsec} (English-centric), \textsc{fs} (seen directions), \textsc{fs+ps+un} (all directions), evaluated on \textit{fully supervised} (de/en/ko/nl/ru/zh pairs), \textit{unsupervised} (cs/is/ja/pl/sv/uk pairs), and \textit{partially supervised} (combining supervised and unsupervised) language pairs.
    Numbers above bars show mean scores.
    Training on more diverse sets improves \textit{all} categories, with \textsc{fs+ps+un} achieving best results even for fully supervised pairs.
    \textsc{fs} substantially reduces off-target rates for unsupervised directions compared to \textsc{base} and \textsc{fsec}, despite these pairs being \textit{absent} from its fine-tuning data.}}    
    \label{fig:comet-strict-off-target-7b-nllb}
\end{figure*}

To investigate whether the performance benefits observed with increased language diversity could be achieved through explicit regularization techniques, we conduct additional experiments using stronger regularization on models with limited language diversity.
If increased language diversity primarily functions as a form of regularization, we hypothesize that similar improvements could be obtained by directly increasing regularization strength in less diverse models.

All our previous experiments use the AdamW optimizer with weight decay set to 0.01 and gradient clipping at 1.0.
This aligns with common practices in LLM fine-tuning, where dropout \citep{srivastava_dropout_2014} is rarely employed (neither the \textsc{Llama} nor \textsc{Tower} papers mention dropout, though both use weight decay).
Notably, AdamW applies weight decay directly to the weights rather than through gradients, decoupling it from the learning rate.

We tested this hypothesis by fine-tuning the \textsc{fs} setup with increased weight decay values of 0.05 and 0.10 (compared to our standard 0.01).
We chose the \textsc{fs} setup to examine whether stronger regularization would induce better cross-lingual transfer to partially supervised and unsupervised directions, potentially mimicking the benefits observed in the more diverse \textsc{fs+ps+un} model.

Figure \ref{fig:regularization} shows the \textsc{COMET-strict} scores comparing \textsc{fs} models with weight decay values of 0.01 (standard) and 0.10.\footnote{The results for weight decay at 0.05 were very similar to 0.10 and are omitted for clarity.}
Increasing the regularization strength has minimal impact on translation performance across all language categories.
For fully supervised directions, both models achieved identical mean scores (0.875).
For partially supervised directions, the difference was negligible (0.699 vs. 0.697).
For unsupervised directions, the model with stronger regularization actually performed slightly worse (0.483 vs. 0.490).

We further explored alternative regularization approaches by implementing LoRA \citep{hu2022lora} with rank 64, which constrains fine-tuning to a low-dimensional subspace.
This parameter-efficient tuning method can be considered a form of regularization as it restricts model updates to a much smaller parameter space than full fine-tuning, potentially preventing overfitting.
Results from LoRA experiments align with our weight decay findings: performance for fully and partially supervised directions remained comparable to full fine-tuning with standard regularization, while unsupervised directions showed slight degradation.

These experiments demonstrate that our initial weight decay value of 0.01 already provides an appropriate balance between overfitting prevention and model flexibility.
More importantly, they confirm that the cross-lingual transfer benefits observed in more diverse models cannot be replicated merely by increasing explicit regularization in less diverse models.
The language diversity benefits we observe go beyond simple explicit regularization effects, providing specialized cross-lingual knowledge transfer.
Our findings align with \citet{aharoni_massively_2019}, who suggest that multilingualism provides benefits beyond what can be achieved through explicit regularization methods.

\begin{figure*}[ht]
    \centering
    \includegraphics[width=.495\linewidth]{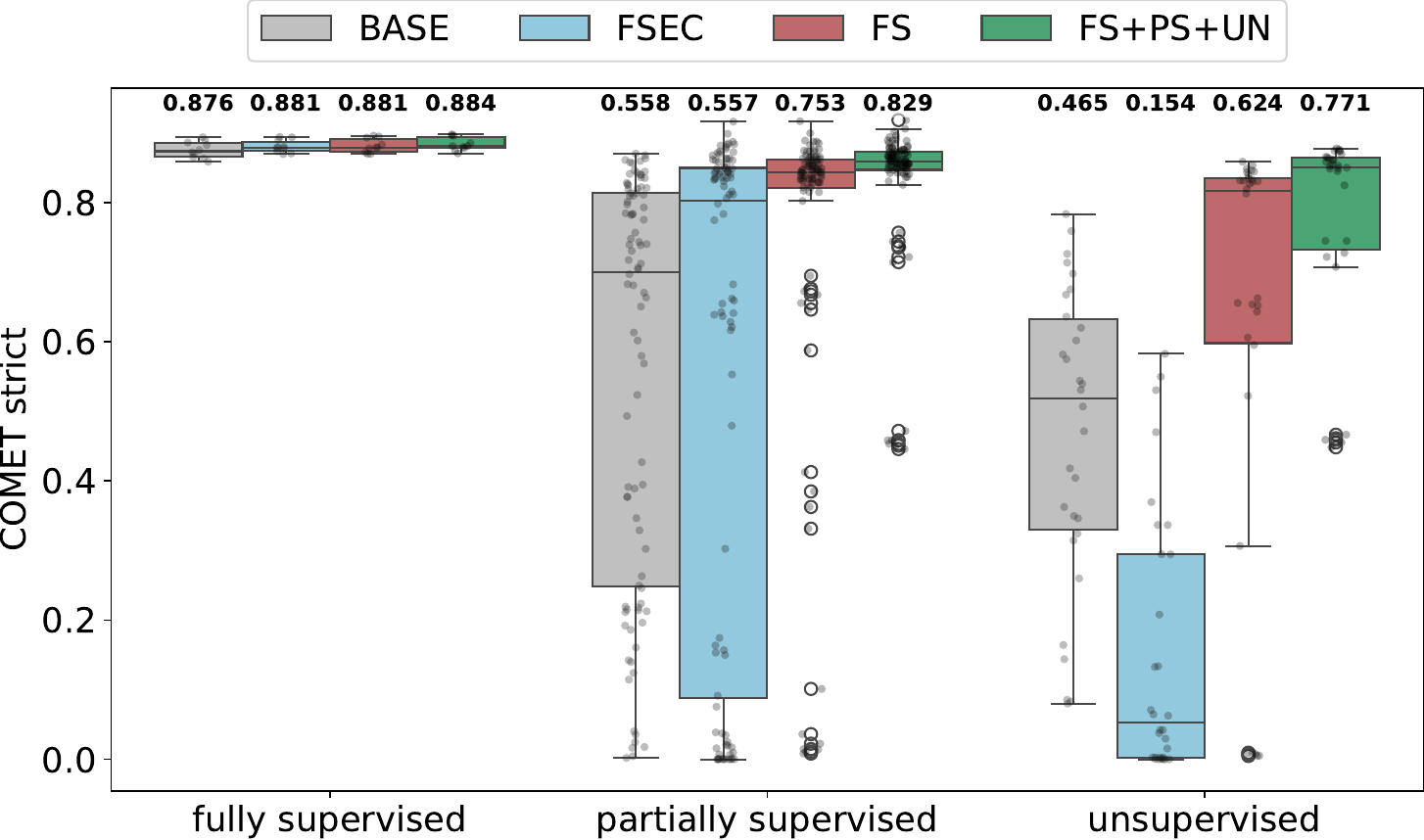}
    \includegraphics[width=.495\linewidth]{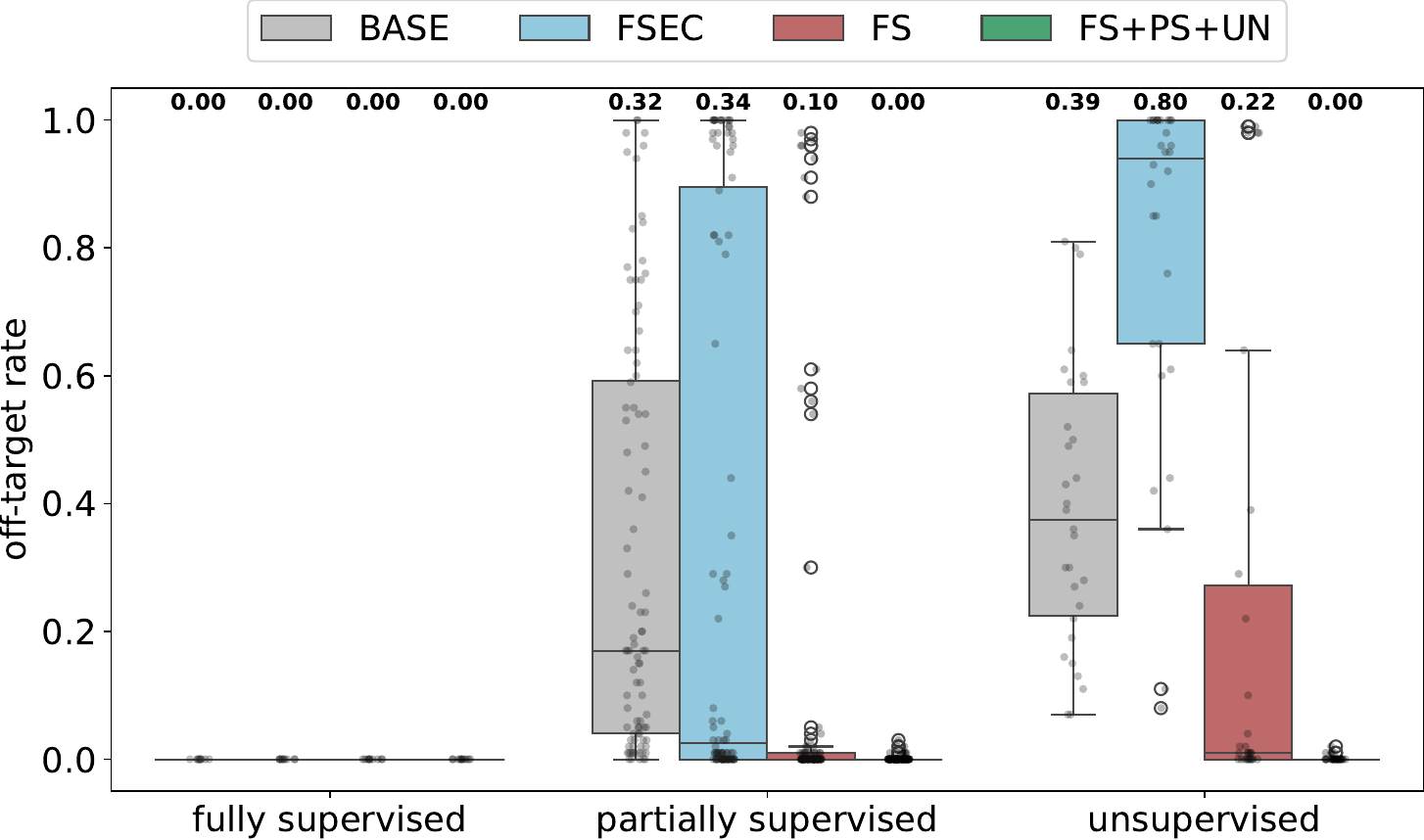}
    \caption{\small{\textsc{COMET-strict} scores for 13B models: \textsc{base} (no fine-tuning), \textsc{fsec} (English-centric), \textsc{fs} (seen directions), \textsc{fs+ps+un} (all directions), evaluated on \textit{fully supervised} (de/en/ko/nl/ru/zh pairs), \textit{unsupervised} (cs/is/ja/pl/sv/uk pairs), and \textit{partially supervised} (combining supervised and unsupervised) language pairs.
    Numbers above bars show mean scores.
    Training on more diverse sets improves \textit{all} categories, with \textsc{fs+ps+un} achieving best results even for fully supervised pairs.
    \textsc{fs} substantially reduces off-target rates for unsupervised directions compared to \textsc{base} and \textsc{fsec}, despite these pairs being \textit{absent} from its fine-tuning data.}}
    \label{fig:comet-strict-off-target-13b}
\end{figure*}

\subsection{Results not due to multi-parallel data}
\label{app:multi_parallel}

To verify our findings are not artifacts of using multi-parallel data, we constructed a non-multi-parallel dataset from the \textsc{NLLB} corpus \citep{nllb_team_no_2022}.
We maintained the same 132 language directions as in our main experiments but eliminated the multi-parallel property
Following \citet{koehn-2024-neural}, we extract examples with \textsc{LASER} \citep{artetxe_massively_2019} scores above 1.05.
We then removed sentences that appeared in multiple language pairs and sampled the remaining data to ensure exactly 2,000 examples per direction, creating a completely non-multi-parallel dataset of equivalent size to our NTREX experiments.

Figure \ref{fig:comet-strict-off-target-7b-nllb} shows \textsc{COMET-strict} (left) and off-target (right) results from experiments conducted using the filtered \textsc{NLLB} dataset rather than \textsc{NTREX}, allowing us to verify that our findings are not artifacts of using multi-parallel data.

The results demonstrate that our core finding—increased language diversity during fine-tuning leads to better performance—holds when using non-multi-parallel data as well.
The \textsc{fs+ps+un} model still achieves the highest \textsc{COMET-strict} scores across all language categories, including for fully supervised language pairs. This confirms that the benefits of diverse fine-tuning extend beyond the multi-parallel setting described in our main experiments.

When comparing performance between models fine-tuned on \textsc{NLLB} versus \textsc{NTREX} data, we observe identical ranking patterns across different fine-tuning setups, though the \textsc{NTREX}-trained models show slightly better overall performance.
This marginal improvement is likely attributable to \textsc{NTREX}'s higher data quality, as it consists of professionally translated content specifically designed for machine translation evaluation.

\subsection{Invariance to model scale}
\label{app:scale_invariance}
Figure \ref{fig:comet-strict-off-target-13b} demonstrates that our findings about language diversity benefits persist when scaling to 13B parameters.

For translation quality (Figure \ref{fig:comet-strict-off-target-13b}, left), the most diverse setup (\textsc{fs+ps+un}) consistently achieves the best results across all language categories, including fully supervised pairs.
While most 13B models show higher scores than their 7B counterparts (Figure \ref{fig:comet-strict-off-target-7b}, left), the \textsc{fsec} model unexpectedly performs worse than \textsc{base} in partially supervised and unsupervised settings (0.557 vs 0.558 and 0.154 vs 0.465), unlike in the 7B configuration where \textsc{fsec} outperformed \textsc{base}.

For off-target rates (Figure \ref{fig:comet-strict-off-target-13b}, right), the most diverse setup again eliminates off-target translations completely.
No model produces off-target translations for fully supervised pairs.
The \textsc{fsec} 13B model shows substantially worse performance for partially supervised (0.34) and unsupervised (0.80) pairs compared to its 7B version (Figure \ref{fig:comet-strict-off-target-7b}, right).
Though \textsc{base} and \textsc{fs} 13B models show improved off-target rates compared to 7B, the problem remains significant (\textsc{base}: 39\% for unsupervised, \textsc{fs}: 22\%).

The decrease in performance for the \textsc{fsec} 13B model can likely be attributed to overfitting to the limited English-centric training data.

These results confirm that language diversity benefits during fine-tuning are robust across model scales, consistently improving both translation quality and target language fidelity.

\end{document}